\documentclass{article}
\usepackage{iclr2027_conference,times}
\usepackage[T1]{fontenc}
\usepackage[utf8]{inputenc}
\usepackage{microtype}
\usepackage{inconsolata}
\usepackage{graphicx}
\usepackage{hyperref}
\usepackage{url}
\usepackage{booktabs}
\usepackage{amsmath,amsfonts,amssymb}
\usepackage{subcaption}
\usepackage[ruled,vlined]{algorithm2e}
\usepackage[table]{xcolor}
\definecolor{decisiondarkred}{HTML}{8B0000}

\usepackage{wrapfig}
\usepackage{enumitem}

\iclrfinalcopy % arXiv version: show author block.
\author{
 \textbf{Yaopei Zeng \textsuperscript{1,3 \footnote{}}},
 \textbf{Congchao Wang \textsuperscript{2}},
 \textbf{Blake JianHang Chen \textsuperscript{3}},
 \textbf{Lu Lin \textsuperscript{1}}
\\
 \textsuperscript{1}Pennsylvania State University,
 \textsuperscript{2}Amazon AGI,
 \textsuperscript{3}Google
 }

\newsavebox{\revtablebox}
\newcommand{\revfit}[1]{\sbox{\revtablebox}{#1}\ifdim\wd\revtablebox>\linewidth\resizebox{\linewidth}{!}{\usebox{\revtablebox}}\else\usebox{\revtablebox}\fi}
\newcommand{\method}{ReLope}
\newcommand{\attn}{Attention Probe}

\title{ReLope: From Hidden-State Probing to a Decision Module for Multimodal LLM Routing}
\begin{document}
\maketitle
% arXiv version: remove the ICLR running header set by the style file.
\lhead{}
\renewcommand{\headrulewidth}{0pt}

\begin{abstract}
Routing balances performance and cost in hybrid systems by escalating selected queries from a lightweight model to a powerful but expensive one.  Hidden-state probes provide an effective routing signal by predicting the small model's correctness from hidden states it already computes. Although effective in text-only settings, their behavior on multimodal LLMs (MLLMs) is less understood. We find that correctness is harder to predict from hidden states when questions are paired with images rather than with captions of those images. Two training-free dependence measures, HSIC and label CKA, show the same pattern. We first investigate whether changing hidden states used by the probe improves correctness prediction for MLLMs. We propose \attn{}, which learns an attention query to pool frozen token states into a routing feature and improves over the standard probe. However, pooling only recombines hidden states computed for answer generation, not for deciding whether the answer should be trusted. We therefore build a decision layer inside the existing MLLM and train it for the routing decision. The resulting router, \method{} (KL-Regularized LoRA Probe), adapts a separate copy of one MLLM layer with LoRA and regularizes it with a KL-penalized stochastic bottleneck. This decision branch shares the frozen lower layers and leaves the model's answer-generation path unchanged. \method{} attains the highest correctness prediction AUC on all five benchmarks and three backbones we evaluate and improves the accuracy versus cost trade-off at negligible overhead.
These results suggest a new perspective and a practical technical path on routing: rather than reading decisions off states built for generation, routing decisions can be learned by a dedicated decision layer inside the model.
Code: \href{https://github.com/Spinozaaa/ReLope}{github.com/Spinozaaa/ReLope}.
\end{abstract}

% However, it changes only how existing states are combined. We therefore introduce \method{} (KL-Regularized LoRA Probe), which directly learns \emph{what} to encode for correctness prediction. It inserts LoRA adapters into a copy of one layer in a routing branch that leaves the answer path frozen, then compresses the adapted state through a stochastic bottleneck, so that correctness supervision shapes the adapted hidden state while its capacity to encode input-specific detail stays bounded.

\ificlrfinal
\begingroup
\renewcommand{\thefootnote}{*}
\footnotetext{Work done during the internship at Google. Correspondence to: Lu Lin \href{mailto:lulin@psu.edu}{<lulin@psu.edu>}}
\endgroup
\fi

\section{Introduction}
\label{sec:intro}
Routing combines the efficiency of a small language model with the capability of a larger one by escalating only selected queries to the larger model~\citep{chen2023frugalgpt,ding2024hybrid,chen2025harnessing}. Hidden-state probing offers an efficient routing signal: a lightweight classifier predicts whether the small model's answer is correct from its hidden states \citep{kadavath2022language,azaria2023internal,chuang2025confident}. Because these hidden states contain information predictive of correctness \citep{orgad2025llms,chen2024inside}, probes route effectively in text-only settings without extra generations or an additional model call to estimate confidence.

It is less clear whether this approach transfers equally well to multimodal LLMs (MLLMs). Prior work shows that vision-language models are often worse calibrated on visual inputs than on semantically equivalent text \citep{xuan2025seeing, zou2025understanding}, and that internal probes transferred across medical VQA domains can perform near chance \citep{khanmohammadi2026calibrated}. These findings raise concerns about multimodal reliability, but do not directly address how effectively lightweight routing probes can predict correctness from frozen MLLM hidden states.

We find that correctness is harder to predict from hidden states when a question comes with an image than when it comes with a caption of that image. 
% To compare these conditions, we keep the questions fixed and replace each image with its caption. On A-OKVQA, this replacement lowers the small model's answer accuracy but increases probe AUC (Table~\ref{tab:motivation}). 
Table~\ref{tab:motivation} shows this on A-OKVQA, where we keep each question fixed and replace its image with a caption.
% Since probe AUC also depends on the classifier and its training, we further examine the dependence between hidden states and correctness labels using HSIC \citep{gretton2005measuring} and label CKA \citep{kornblith2019similarity}, neither of which requires classifier training. Both measures indicate stronger dependence in the caption-only condition, consistent with the higher probe AUC. The same pattern appears on two other benchmarks (Appendix~\ref{app:e05}). This suggests that the gap in correctness prediction reflects differences in the hidden states used as probe inputs, rather than arising solely from the probe's architecture or parameters.
Using the textual caption, the small model answers less accurately (68.73\% vs.\ 78.78\%), yet its correctness is easier to predict (probe AUC 85.25 vs.\ 81.94). The gap is not specific to the trained probe: HSIC \citep{gretton2005measuring} and label CKA \citep{kornblith2019similarity}, which measure the dependence between hidden states and correctness labels without training a classifier, are also higher with captions. The same pattern holds on two other benchmarks (Appendix~\ref{app:e05}). With image inputs, the correctness signal is therefore harder to extract from the hidden states themselves.
% Routing quality and answer quality therefore do not move together, and the difficulty is not an artifact of the probe. Section~\ref{sec:preliminaries} gives the input conditions and the protocol.

\begin{table}[t]
\caption{\textbf{Correctness becomes harder to probe with multimodal input.} Three input conditions on A-OKVQA with Qwen2.5-VL-7B-Instruct, differing in evidence source and the presence of image tokens. See Appendix~\ref{app:input_examples} for examples. HSIC and label CKA are dependence measures between hidden states and correctness labels that require no classifier training.}
\label{tab:motivation}
\vspace{-10pt}
\centering\small\setlength{\tabcolsep}{3pt}
\revfit{\begin{tabular}{lcccccc}
\toprule
Input & Evidence & Image tokens & LLM Acc.\ (\%) & Probe AUC & HSIC ($\times10^{-3}$) & Label CKA\\
\midrule
Original input & image & \checkmark & \textbf{78.78} & $81.94\pm0.58$ & 2.115 & 0.1123\\
Caption only & text & -- & 68.73 & $\mathbf{85.25}\pm0.52$ & \textbf{9.842} & \textbf{0.2171}\\
Caption + random image & text & \checkmark & 68.30 & $84.73\pm1.08$ & 8.364 & 0.1952\\
\bottomrule
\end{tabular}}
\vspace{-15pt}
\end{table}

Motivated by this finding, we ask whether changing the hidden state the probe reads can improve correctness prediction. 
A standard probe reads the last-token hidden state at a selected layer. The MLLM produces this state for answer generation, not targeted to expose whether its answer is correct, and with multimodal inputs, the state must encode both visual and textual content. We therefore first ask whether learning how to combine token hidden states helps. We propose \attn{}, which uses a learned attention query to weigh the frozen token hidden states and passes their weighted sum to the correctness classifier. The query and classifier are trained jointly on correctness labels, while the MLLM remains frozen. On A-OKVQA, \attn{} reduces the AUC gap between image and caption inputs from 3.31 to 1.25 points (Section~\ref{sec:ablation}), suggesting that how the hidden states are pooled contributes substantially to the observed gap.

% Although learned pooling improves correctness prediction, it changes only how existing hidden states are combined. We therefore introduce \method{} (KL-Regularized LoRA Probe), which directly learns \emph{what} the hidden states encode for correctness prediction. \method{} inserts LoRA adapters into a separate copy of the selected layer \citep{hu2022lora}, forming a routing branch while leaving the answer-generating model frozen. This allows correctness supervision to shape the routing hidden states without changing the small model's answers or their correctness labels. However, adaptation also increases the risk of fitting input-specific details that do not generalize. We therefore pass the adapted last-token hidden state through a stochastic bottleneck with a KL penalty \citep{alemi2016deep}. Together, the prediction loss and KL regularization encourage the routing feature to retain information useful for correctness prediction while limiting unnecessary input-specific detail. Neither method requires captions, extra generations, or a large-model call before the routing decision.

However, learned pooling only changes how frozen hidden states are combined.
The states themselves are still shaped for answer generation, not for deciding whether that answer should be trusted. Prior work finds that correctness prediction improves when the model itself is trained for it rather than probed frozen~\citep{kadavath2022language, kapoor2024large}, and Jev~\citep{almeida2026jev} makes a related argument at the system level: LLMs are optimized to generate text, not to make calibrated decisions. Sharing this principle, rather than reading the routing decision off these states, we build a decision layer inside the existing MLLM and train it for the routing decision. We call the resulting router \method{} (KL-Regularized LoRA Probe). It inserts LoRA~\citep{hu2022lora} adapters into a separate copy of the selected layer, and trains this copy for correctness classification. The copy reuses the model's frozen lower layers and forms a routing branch, leaving the answer-generating model unchanged. Correctness supervision can thus shape the routing hidden states without changing the small model's answers or their correctness labels. Adaptation, however, also raises the risk of fitting input-specific details that do not generalize. We therefore pass the adapted last-token hidden state through a stochastic bottleneck with a KL penalty \citep{alemi2016deep}. Together, the prediction loss and the KL penalty encourage the routing feature to retain information useful for correctness prediction while discarding unnecessary input-specific detail. Neither method requires captions, extra generations, or a large-model call before the routing decision.

Across five multimodal benchmarks and three backbones, \method{} achieves the highest mean correctness prediction AUC among the evaluated methods in all 15 dataset--backbone pairs. It also improves the hybrid system's accuracy--cost trade-off with low inference overhead. Our contributions are:
\begin{itemize}[leftmargin=*]

\item An empirical analysis showing that correctness is harder to predict from MLLM hidden states under image inputs than under caption inputs, together with evidence that learned pooling substantially narrows this gap.
\item \attn{}, which learns how to pool frozen hidden states for correctness prediction, and \method{}, which combines adaptation of one layer with a KL-regularized stochastic bottleneck while preserving the small model's answers.
\item An evaluation across five benchmarks and three backbones covering correctness prediction, hybrid accuracy and cost, inference overhead, component ablations, robustness, and cross-dataset transfer.
\end{itemize}

\section{Related Work}
\label{sec:related}
\textbf{Model selection and routing.}
Input-based routers choose a model from the input query alone, using routers trained on benchmark evaluation results, reward signals, preference data, or learned taxonomies \citep{shnitzer2023large,lu2023routing,ong2024routellm,mohammadshahi2024routoo,shah2025select,panda2025adaptive}. Eagle ranks models without training a router \citep{zhao2024eagle}, and Router-R1 chains several model calls and aggregates their answers \citep{zhang2025router}. Universal routers treat routing and cascading as one problem \citep{jitkrittum2025universal}. We study the two-model decision after the small model has run, using features it has already computed.

\textbf{Cascades and probes.}
Cascades defer selected queries after a smaller model has answered \citep{chen2023frugalgpt,yue2023large}, relating routing to selective prediction \citep{chow1970optimum,geifman2017selective}. Deferral signals include output probabilities, token-level uncertainty and answer agreement \citep{huang2023look,gupta2024language}, output embeddings, and hidden-state probes \citep{kadavath2022language,azaria2023internal,chuang2025confident}. Their reliability for factual correctness estimation has also been studied directly \citep{mahaut2024factual}. Probe performance depends on both the hidden state and the classifier, so limited decoding accuracy does not by itself establish the absence of information \citep{alain2016understanding,belinkov2022probing,hewitt2019designing}. We study readout learning and supervised adaptation of the hidden states for multimodal routing. 

Closest to our design, \citet{kapoor2024large} find that probes on frozen features and prompting give weaker uncertainty estimates than fine-tuning on a small set of graded answers. Their estimator is text-only and works by prompting the fine-tuned model with an uncertainty query. \method{} instead adapts a copy of a single layer on top of the frozen lower layers without interfering with the model itself. At the system level, Jev \citep{almeida2026jev} argues that decision tasks call for models trained for calibrated decisions rather than for text generation. \method{} applies a similar principle within a lightweight one-layer branch of an existing MLLM. 

\textbf{Multimodal confidence and hidden states.}
MLLMs form hidden states from both visual and textual tokens \citep{yin2024survey}, while some benchmark questions remain answerable without the image \citep{chen2024mmstar}. Reliability therefore requires more than high answer accuracy. \citet{xuan2025seeing} show that verbalized calibration is often worse with visual inputs than with semantically equivalent text. \citet{khanmohammadi2026calibrated} find that confidence estimation remains unreliable across medical VQA domains when probes trained on natural images are transferred without adaptation. VIB-Probe \citep{zhang2026vib} uses a variational bottleneck on attention-head outputs for hallucination detection and generation-time intervention. \method{} instead predicts correctness from prefill hidden states for multimodal routing, jointly training an adapted layer and a stochastic bottleneck in a separate branch while keeping the answer-generating model frozen. 

\begin{figure}[t]
\centering
% Correct the label in the imported figure without changing its layout.
\begingroup
\setlength{\unitlength}{1bp}
\resizebox{\linewidth}{!}{%
\begin{picture}(703.4,182.28)
\put(0,0){\includegraphics[width=703.4bp]{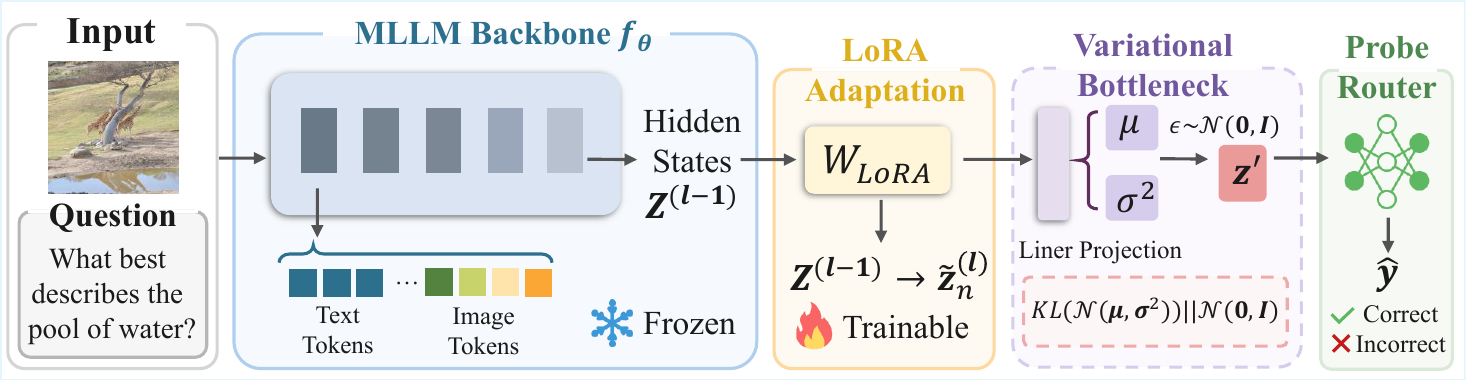}}
\put(488.5,54.8){{\color[RGB]{250,248,253}\rule{84bp}{15bp}}}
\put(488.92,58.28){{\normalfont\fontsize{12bp}{14bp}\selectfont Linear Projection}}
\end{picture}%
}
\endgroup
\caption{\textbf{\method{} overview.} Token states $\mathbf{Z}^{(l-1)}$ of the frozen small model enter a copy of layer $l$ with LoRA adapters. The adapted last-token state is projected to a Gaussian $(\boldsymbol\mu,\boldsymbol\sigma^2)$, a sample from which is classified for correctness. A KL penalty to a standard normal prior regularizes the latent feature. LoRA weights, projection heads, and classifier are trained together with the routing loss.}
\label{fig:pipeline}
\vspace{-15pt}
\end{figure}

\section{Preliminaries}
\label{sec:preliminaries}
\textbf{Problem statement.}
We consider a hybrid MLLM system consisting of a small, cheap model and a larger, more capable model. Given a multimodal input $\mathbf{x}$ containing a text prompt and an image, the small model produces an answer $h_s(\mathbf{x})$. The binary label $y$ indicates whether the small model's answer is correct. A router assigns a correctness score $s(\mathbf{x})$, with higher values indicating greater confidence in the small model's answer. The system retains that answer when $s(\mathbf{x})\geq\tau$ and otherwise escalates the query to the large model. The threshold $\tau$ controls the escalation rate. 

\textbf{Hidden-state probe.} At layer $l$ the model produces token hidden states $\mathbf{Z}^{(l)}=(\mathbf{z}_1^{(l)},\ldots,\mathbf{z}_n^{(l)})\in\mathbb{R}^{n\times d}$. The standard probe \citep{alain2016understanding,chuang2025confident} reads the last-token state and trains a classifier $g_\phi$ with binary cross-entropy (BCE) against $y$: $ s_{\mathrm{P}}(\mathbf{x})=\operatorname{sigmoid}\bigl(g_\phi(\mathbf{z}_n^{(l)})\bigr).
 \label{eq:standard}$
The backbone stays frozen, and the last-token readout is fixed by the probe design. The classifier learns to predict correctness from this hidden state without changing how it is formed. Its performance reflects both the available signal and how readily the chosen classifier can extract it.

\textbf{Correctness prediction under multimodal input.}
We compare image and caption inputs to examine how the input modality affects hidden-state probing. Using Qwen2.5-VL-7B-Instruct on the same A-OKVQA questions, we consider three conditions: the original image, a caption replacing the image, and the caption paired with an unrelated image. The third condition tests whether adding image tokens affects correctness prediction when the relevant evidence is still provided as text. Answers and correctness labels are recomputed for each condition.

As shown in Table~\ref{tab:motivation}, replacing images with captions increases probe AUC from 81.94 to 85.25 while lowering answer accuracy from 78.78\% to 68.73\%. Since probe AUC also depends on the classifier and its training, we additionally measure dependence between hidden states and correctness labels using HSIC \citep{gretton2005measuring} and label CKA \citep{kornblith2019similarity}. Neither measure requires classifier training, and both yield higher values in the caption-only condition, consistent with the probe results. The same image-versus-caption pattern appears on two additional benchmarks (Appendix~\ref{app:e05}). These comparisons are descriptive, as changing the input also changes the answers and correctness labels.

\section{Learning the Routing Feature}
\label{sec:method}

Table~\ref{tab:motivation} shows that correctness is harder to probe under multimodal input, and the classifier-free measures place the difficulty outside the classifier. What is left is the hidden state the probe reads directly from the MLLM, the last-token state of layer $l$. Motivated by these observations, we investigate two ways to learn the hidden state used for correctness prediction. Both methods use the frozen token hidden states $\mathbf{Z}^{(l-1)}$ as their starting point. \attn{} learns an attention query that pools these hidden states into an input for the correctness classifier. \method{} instead passes them through a separately adapted copy of layer $l$, then applies a stochastic bottleneck to the adapted last-token hidden state. Both methods use the same classifier architecture $g_\phi$, with BCE supervising correctness prediction; \method{} additionally uses KL regularization. Figure~\ref{fig:pipeline} illustrates \method{}, and Algorithm~\ref{alg:routing} summarizes training for both methods.

\begin{algorithm}[htbp]
\small
\caption{Training \attn{} or \method{} (separate runs)}
\label{alg:routing}
\KwIn{Correctness dataset $\mathcal D=\{(\mathbf x,y)\}$; frozen backbone $f_\theta$; read layer $l$; chosen router}
\For{each training epoch}{
\For{each minibatch $(\mathbf x,y)$}{
 $\mathbf Z^{(l-1)}\gets f_\theta^{(<l)}(\mathbf x)$\;
 \eIf{the chosen router is \attn{}}{
  Compute $\boldsymbol\omega$ and $\mathbf z_{\mathrm A}$ with Eq.~\ref{eq:attention}\;
  $\mathcal L\gets\operatorname{BCE}(y,\operatorname{sigmoid}(g_\phi(\mathbf z_{\mathrm A})))$; update $(\mathbf q,\phi)$\;
 }{
  $\tilde{\mathbf Z}^{(l)}\gets f^{(l)}_{\theta,\theta_{\mathrm{LoRA}}}(\mathbf Z^{(l-1)})$; $\mathbf z_{\mathrm L}\gets\tilde{\mathbf z}^{(l)}_n$\;
  $(\boldsymbol\mu,\log\boldsymbol\sigma^2)\gets\psi(\mathbf z_{\mathrm L})$; sample $\mathbf z_{\mathrm B}$ with Eq.~\ref{eq:bottleneck}\;
  Compute $\mathcal L_{\mathrm R}$ with Eq.~\ref{eq:relope}; update $(\theta_{\mathrm{LoRA}},\psi,\phi)$\;
 }
}
}
\tcp{At inference \method{} uses $\mathbf z_{\mathrm B}=\boldsymbol\mu$.}
\vspace{-15pt}
\end{algorithm}

\subsection{\attn{}: learn where to read}
\label{sec:reorganize}
\textbf{Motivation.}
The hidden state the probe reads is the last-token state of layer $l$, the strongest fixed readout in our sweep over positions and layers (Section~\ref{sec:ablation}). However, the MLLM produces this hidden state for answer generation, without an explicit objective to make answer correctness easy to predict. We therefore investigate whether learning how to pool token hidden states specifically for correctness prediction improves probe performance.

\textbf{Design.}
\attn{} applies attention pooling to the frozen token hidden states at layer $l-1$. A learned query $\mathbf q\in\mathbb R^d$ assigns a score to each token hidden state. A softmax converts these scores into weights, and the resulting weighted sum is passed to the correctness classifier:
\begin{equation}
\omega_i=\frac{\exp(\mathbf{q}^{\top}\mathbf{z}_i^{(l-1)}/\sqrt{d})}
 {\sum_{j=1}^{n}\exp(\mathbf{q}^{\top}\mathbf{z}_j^{(l-1)}/\sqrt{d})},
 \qquad
 \mathbf{z}_{\mathrm A}=\sum_{i=1}^{n}\omega_i\mathbf{z}_i^{(l-1)},
 \qquad
 s_{\mathrm A}(\mathbf{x})=\operatorname{sigmoid}\bigl(g_\phi(\mathbf{z}_{\mathrm A})\bigr).
 \label{eq:attention}
\end{equation}

The sum includes all valid text and image tokens, allowing the probe to draw on hidden states from both modalities and across token positions.
To keep the pooling mechanism lightweight, we use a single learned query $\mathbf q$ shared across samples and token positions. This introduces only $d$ trainable pooling parameters and applies the same scoring rule to every token, while the resulting weights adapt to each input's hidden states. We scale the dot-product scores by $1/\sqrt d$ to reduce the risk of softmax saturation. Initializing $\mathbf q$ near zero gives approximately uniform weights, so the probe starts from a readout close to mean pooling and learns to emphasize different token hidden states through correctness supervision.

Training updates only the query $\mathbf q$ and classifier parameters $\phi$. The MLLM and its hidden states remain frozen. The classifier architecture matches that of the standard probe, allowing us to compare a fixed hidden state with learned attention pooling while keeping the backbone unchanged. Section~\ref{sec:ablation} examines how learned pooling affects correctness prediction under image and caption inputs.

\subsection{\method{}: learn what to encode}
\label{sec:kl-lora}
\textbf{Motivation.}
% \attn{} learns how to combine existing hidden states, but leaves the computation of those hidden states unchanged. We next investigate whether adapting the hidden states themselves under correctness supervision can further improve prediction. \method{} learns \emph{what} to encode for this task while preserving the small model's answers.
\attn{} learns how to combine frozen hidden states, but the states themselves are still computed for answer generation. Sharing the principle of training models specifically for decisions~\citep{almeida2026jev, kapoor2024large}, we build a decision layer inside the existing MLLM and train it for the routing decision, so that correctness supervision shapes what the routing feature encodes.
To keep this cheap, the decision layer is a separate copy of layer $l$ with LoRA adapters that reuses the frozen lower layers, leaving the model's original answer path unchanged.

Adaptation must preserve the small model's answers and add little computation. Prior work shows that visual tokens can contain redundancy and information irrelevant to the question \citep{yang2025visionzip, cao2023pumer}. Motivated by these findings, we hypothesize that retaining information relevant to correctness while limiting unrelated visual details can make correctness easier to predict from hidden states. \method{} therefore combines single-layer adaptation with a KL-regularized stochastic bottleneck. Correctness supervision guides the adapted hidden state toward correctness prediction, while the bottleneck limits how much input-specific information the routing feature retains.
% \yp{Prior work identifies redundancy and question-irrelevant information in visual token sequences \citep{yang2025visionzip, cao2023pumer}. We therefore hypothesize that retaining information relevant to correctness while limiting unrelated visual details can improve prediction. In \method{}, correctness supervision shapes the adapted hidden state, while a KL-regularized stochastic bottleneck limits the input-specific information retained in the routing feature.}

\textbf{Layer adaptation.}
We insert LoRA adapters into a separate copy of layer $l$ while freezing its original weights. For an adapted weight matrix, the update is $\Delta\mathbf W=(\alpha/r)\mathbf B\mathbf A$ with rank at most $r$ \citep{hu2022lora}. The adapted layer maps $\mathbf Z^{(l-1)}$ to $\tilde{\mathbf Z}^{(l)}$, from which we take the last-token state $\mathbf z_{\mathrm L}=\tilde{\mathbf z}^{(l)}_n$. The routing branch shares the lower layers with the original answer path and adds one layer computation. Retaining the last-token interface allows us to evaluate adaptation of the hidden states without simultaneously changing the token selection rule.

\textbf{Stochastic bottleneck.}
Our HSIC and label CKA analysis examines the dependence between frozen hidden states and correctness labels. This motivates learning a routing feature that retains information predictive of correctness. To regularize this learning process, we adopt the information bottleneck principle: preserve information useful for correctness prediction while limiting how much information the feature retains from the adapted hidden state \citep{tishby2015deep,alemi2016deep}. We implement this trade-off with a stochastic bottleneck trained using correctness supervision and a KL penalty. Specifically, two linear heads parameterized by $\psi$ map $\mathbf z_{\mathrm L}$ to $\boldsymbol\mu$ and $\log\boldsymbol\sigma^2$, defining a Gaussian distribution:
\begin{equation}
 q_\psi(\mathbf z_{\mathrm B}\mid\mathbf z_{\mathrm L})
 =\mathcal N(\boldsymbol\mu,\operatorname{diag}(\boldsymbol\sigma^2)),
 \qquad
 \mathbf z_{\mathrm B}=\boldsymbol\mu+\boldsymbol\sigma\odot\boldsymbol\epsilon,
 \quad\boldsymbol\epsilon\sim\mathcal N(\mathbf0,\mathbf I).
 \label{eq:bottleneck}
\end{equation}
During training, we sample $\mathbf z_{\mathrm B}$ using the reparameterization trick \citep{kingma2014auto}, allowing gradients to propagate through the stochastic bottleneck. At inference, we use the mean $\boldsymbol\mu$ instead of sampling, yielding the deterministic score
$s_{\mathrm R}(\mathbf x)=\operatorname{sigmoid}(g_\phi(\boldsymbol\mu))$.

\textbf{Training objective.} We minimize the correctness loss with a KL penalty toward a fixed prior,
\begin{equation}
\begin{aligned}
\mathcal L_{\mathrm R}=\mathbb E_{(\mathbf x,y)}\Big[&
 \mathbb E_{q_\psi}\!\left[\operatorname{BCE}\bigl(y,\operatorname{sigmoid}(g_\phi(\mathbf z_{\mathrm B}))\bigr)\right]+\beta D_{\mathrm{KL}}\!\left(q_\psi(\mathbf z_{\mathrm B}\mid\mathbf z_{\mathrm L})\,\|\,\mathcal N(\mathbf0,\mathbf I)\right)\Big].
\end{aligned}
\label{eq:relope}
\end{equation}
The BCE term encourages accurate correctness prediction. The expected KL term upper-bounds the mutual information between the adapted hidden state and the stochastic bottleneck output, thereby regularizing how much input-specific information the routing feature retains. The coefficient $\beta$ controls the trade-off between prediction and compression. We jointly optimize the LoRA adapters, projection heads, and classifier, while keeping the answer-generating MLLM frozen.
Section~\ref{sec:transfer} evaluates prediction of clean-input correctness under image perturbations and finds that \method{} has the smallest mean AUC drop among the tested probes.

% \textbf{Inference and routing overhead.}
% Both methods compute a correctness score from the small model's hidden states and compare it with a threshold to make the routing decision. \attn{} adds attention pooling and a classifier. \method{} adds one adapted layer computation, a mean projection head, and a classifier; sampling and the variance head are not needed at inference. Neither method requires extra generated tokens or a large-model call to compute the routing score. 
% Section~\ref{sec:system} reports the measured inference overhead.

\section{Experiments}
\label{sec:experiments}
We ask four questions. Does the learned feature predict correctness better than existing routers? Does that prediction improve a hybrid system, and at what cost? Where does the gain come from? And how far does it carry beyond the training distribution?

% \lu{do not use resizebox, which will make fontsize clearly different from maintext}
\begin{table}[t]
\caption{\textbf{Correctness prediction.} Routing AUC (\%) on five benchmarks and three small backbones, mean $\pm$ standard deviation over three seeds. Bold marks the best mean per column.}
\label{tab:auc_baselines}
\centering
\small\setlength{\tabcolsep}{4pt}
% \resizebox{\linewidth}{!}{%
\begin{tabular}{lccccc}
\toprule
\textbf{Method} & \textbf{MMMU} & \textbf{A-OKVQA} & \textbf{ScienceQA} & \textbf{ChartQA} & \textbf{MathVision} \\
\midrule
\multicolumn{6}{c}{\textit{Qwen2.5-VL-7B-Instruct}} \\
\midrule
MMR-MLP & 63.50 $\pm$ 0.23 & 61.40 $\pm$ 0.57 & 80.05 $\pm$ 0.13 & 71.54 $\pm$ 0.18 & 77.58 $\pm$ 1.19 \\
MMR-CMR & 62.11 $\pm$ 1.15 & 64.77 $\pm$ 2.04 & 82.95 $\pm$ 0.29 & 73.73 $\pm$ 0.27 & 80.25 $\pm$ 2.04 \\
MoT & 66.97 $\pm$ 1.24 & 72.26 $\pm$ 0.62 & 71.83 $\pm$ 0.89 & 76.22 $\pm$ 2.08 & 81.89 $\pm$ 1.64 \\
Cascade Routing & 68.41 $\pm$ 1.14 & 73.13 $\pm$ 0.08 & 76.15 $\pm$ 0.47 & 69.75 $\pm$ 2.21 & 78.20 $\pm$ 1.00 \\
RouteLLM & 71.02 $\pm$ 1.36 & 68.74 $\pm$ 0.65 & 82.78 $\pm$ 0.15 & 75.35 $\pm$ 1.10 & 87.38 $\pm$ 0.88 \\
Post-Hoc Embed & 70.29 $\pm$ 0.03 & 69.97 $\pm$ 0.76 & 75.41 $\pm$ 0.35 & 72.01 $\pm$ 2.24 & 87.73 $\pm$ 1.99 \\
Probe & 76.61 $\pm$ 0.20 & 81.94 $\pm$ 0.58 & 83.65 $\pm$ 0.86 & 81.59 $\pm$ 2.35 & 90.03 $\pm$ 2.05 \\
BEST-Route & 77.51 $\pm$ 0.77 & 80.01 $\pm$ 0.37 & 82.73 $\pm$ 0.69 & 82.51 $\pm$ 2.03 & 90.33 $\pm$ 2.59 \\
\rowcolor{gray!15} Attention Probe (ours) & 78.36 $\pm$ 1.02 & 84.06 $\pm$ 0.65 & 86.11 $\pm$ 0.79 & 85.45 $\pm$ 1.43 & 89.16 $\pm$ 2.51 \\
\rowcolor{gray!15} ReLope (ours) & \textbf{80.79 $\pm$ 0.40} & \textbf{86.24 $\pm$ 0.63} & \textbf{91.78 $\pm$ 0.52} & \textbf{85.94 $\pm$ 0.51} & \textbf{93.56 $\pm$ 0.95} \\
\midrule
\multicolumn{6}{c}{\textit{Gemma3-12B}} \\
\midrule
MMR-MLP & 64.23 $\pm$ 3.89 & 61.62 $\pm$ 0.52 & 84.34 $\pm$ 0.04 & 72.10 $\pm$ 0.04 & 68.71 $\pm$ 0.25 \\
MMR-CMR & 64.49 $\pm$ 1.44 & 64.84 $\pm$ 1.43 & 86.70 $\pm$ 0.21 & 74.05 $\pm$ 0.27 & 72.28 $\pm$ 2.09 \\
MoT & 62.25 $\pm$ 0.47 & 68.11 $\pm$ 0.29 & 72.85 $\pm$ 0.07 & 69.75 $\pm$ 1.44 & 65.66 $\pm$ 0.96 \\
Cascade Routing & 62.47 $\pm$ 0.15 & 64.76 $\pm$ 0.62 & 75.85 $\pm$ 0.70 & 73.08 $\pm$ 0.71 & 67.75 $\pm$ 1.35 \\
RouteLLM & 65.25 $\pm$ 1.39 & 66.11 $\pm$ 0.13 & 78.90 $\pm$ 0.34 & 74.51 $\pm$ 2.39 & 70.37 $\pm$ 2.05 \\
Post-Hoc Embed & 65.57 $\pm$ 0.31 & 71.71 $\pm$ 0.47 & 76.13 $\pm$ 0.60 & 73.51 $\pm$ 1.63 & 68.38 $\pm$ 2.50 \\
Probe & 68.10 $\pm$ 0.46 & 75.10 $\pm$ 0.09 & 80.41 $\pm$ 0.26 & 77.21 $\pm$ 2.47 & 71.71 $\pm$ 1.26 \\
BEST-Route & 69.23 $\pm$ 0.41 & 71.90 $\pm$ 0.62 & 78.85 $\pm$ 1.00 & 75.75 $\pm$ 1.85 & 73.00 $\pm$ 2.99 \\
\rowcolor{gray!15} Attention Probe (ours) & 70.64 $\pm$ 1.46 & 74.49 $\pm$ 0.34 & 85.29 $\pm$ 1.04 & 81.02 $\pm$ 0.75 & 70.39 $\pm$ 2.65 \\
\rowcolor{gray!15} ReLope (ours) & \textbf{76.82 $\pm$ 0.86} & \textbf{85.78 $\pm$ 0.12} & \textbf{88.65 $\pm$ 0.24} & \textbf{87.35 $\pm$ 1.38} & \textbf{81.26 $\pm$ 1.89} \\
\midrule
\multicolumn{6}{c}{\textit{Phi-4-Multimodal-Instruct}} \\
\midrule
MMR-MLP & 56.44 $\pm$ 4.61 & 58.17 $\pm$ 1.18 & 75.07 $\pm$ 1.73 & 60.96 $\pm$ 1.37 & 75.81 $\pm$ 0.28 \\
MMR-CMR & 58.90 $\pm$ 1.44 & 59.33 $\pm$ 1.93 & 77.66 $\pm$ 1.98 & 63.06 $\pm$ 1.13 & 78.74 $\pm$ 0.52 \\
MoT & 64.41 $\pm$ 1.27 & 68.43 $\pm$ 0.15 & 74.08 $\pm$ 0.81 & 66.57 $\pm$ 1.70 & 73.61 $\pm$ 2.16 \\
Cascade Routing & 63.87 $\pm$ 0.30 & 70.11 $\pm$ 0.71 & 75.38 $\pm$ 0.67 & 63.41 $\pm$ 2.12 & 76.70 $\pm$ 1.13 \\
RouteLLM & 66.50 $\pm$ 1.05 & 71.04 $\pm$ 0.62 & 84.20 $\pm$ 0.32 & 70.31 $\pm$ 1.55 & 78.42 $\pm$ 0.94 \\
Post-Hoc Embed & 66.85 $\pm$ 1.00 & 72.75 $\pm$ 0.68 & 77.83 $\pm$ 0.03 & 70.00 $\pm$ 1.91 & 77.18 $\pm$ 1.28 \\
Probe & 69.66 $\pm$ 1.05 & 75.98 $\pm$ 0.58 & 82.00 $\pm$ 0.94 & 73.99 $\pm$ 0.53 & 81.86 $\pm$ 2.75 \\
BEST-Route & 71.55 $\pm$ 0.13 & 75.39 $\pm$ 0.77 & 80.71 $\pm$ 0.76 & 70.24 $\pm$ 2.18 & 80.68 $\pm$ 1.95 \\
\rowcolor{gray!15} Attention Probe (ours) & 72.51 $\pm$ 1.47 & 80.12 $\pm$ 0.69 & 87.20 $\pm$ 0.51 & 78.33 $\pm$ 0.89 & 83.82 $\pm$ 0.87 \\
\rowcolor{gray!15} ReLope (ours) & \textbf{78.64 $\pm$ 0.54} & \textbf{85.89 $\pm$ 0.10} & \textbf{92.55 $\pm$ 0.63} & \textbf{81.64 $\pm$ 0.75} & \textbf{90.09 $\pm$ 2.16} \\
\bottomrule
\end{tabular}
% }
\end{table}

\subsection{Setup}
\label{sec:exp_setup}

\textbf{Datasets and models.}
We evaluate on five multimodal benchmarks: MMMU for general reasoning \citep{yue2024mmmu}, A-OKVQA for knowledge-based visual question answering \citep{schwenk2022okvqa}, the image subset of ScienceQA for science questions \citep{lu2022learn}, ChartQA for chart understanding \citep{masry2022chartqa}, and MathVision for visual mathematics \citep{wang2024measuring}. We use three small backbones: Qwen2.5-VL-7B-Instruct \citep{bai2025qwen2}, Gemma3-12B \citep{team2025gemma}, and Phi-4-Multimodal-Instruct \citep{abouelenin2025phi}. More details are in Appendix~\ref{sec:implementation}.

\textbf{Baselines.}
We compare with eight routers: MoT \citep{yue2023large}, Cascade Routing \citep{chen2023frugalgpt}, RouteLLM \citep{ong2024routellm}, Post-Hoc Embed \citep{gupta2024language}, BEST-Route \citep{ding2025bestroute}, MMR-MLP and MMR-CMR from MMR-Bench \citep{ma2026mmrbench}, and the standard hidden-state probe \citep{chuang2025confident}. These methods use signals ranging from sampled-answer agreement and output statistics to query embeddings and hidden states. We adapt MMR-MLP and MMR-CMR to binary correctness prediction using frozen OpenCLIP ViT-B/32 embeddings. 

\subsection{Correctness prediction}
\label{sec:prediction}
Table~\ref{tab:auc_baselines} shows that frozen hidden states already provide useful routing signals: the standard probe outperforms most output-, sampling-, and preference-based baselines. Building on this baseline, \attn{} learns to combine frozen token hidden states and improves over the standard probe in 12 of the 15 dataset--backbone pairs. \method{} improves over \attn{} in all 15 pairs and achieves the highest mean AUC among the evaluated methods. These results suggest benefits from adapting and regularizing routing hidden states beyond just combining frozen states. Section~\ref{sec:ablation} examines adaptation, the bottleneck, parameter count, and alternative readouts of frozen states.

\subsection{From prediction to a hybrid system}
\label{sec:system}
\begin{figure}[t]
\centering
\begin{subfigure}[t]{.32\linewidth}\centering
\includegraphics[width=\linewidth]{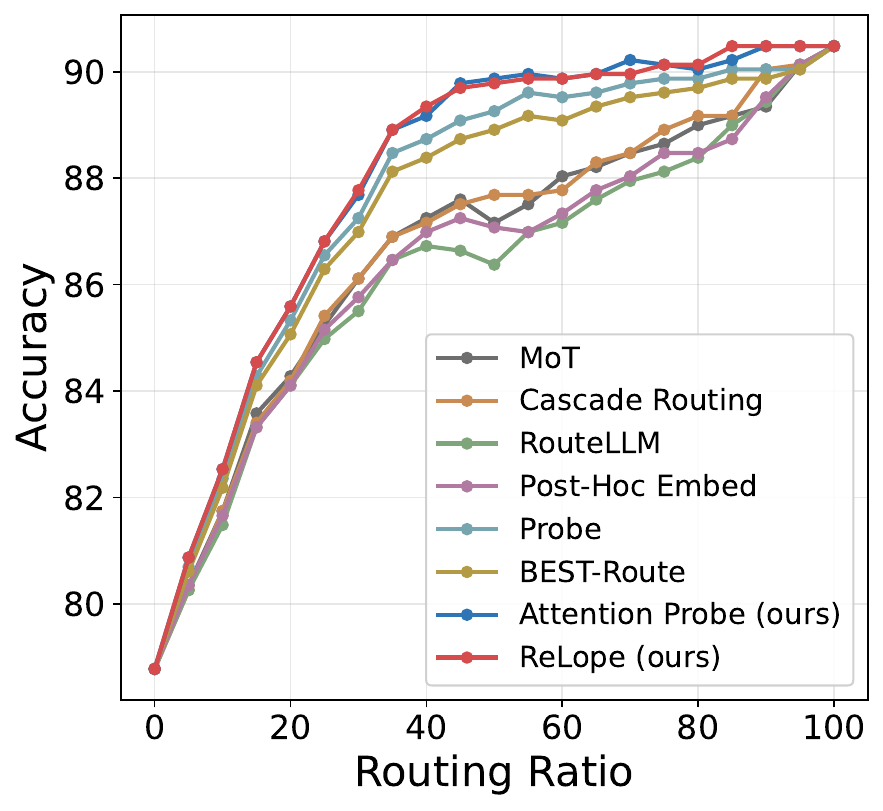}
\caption{A-OKVQA}\end{subfigure}\hfill
\begin{subfigure}[t]{.32\linewidth}\centering
\includegraphics[width=\linewidth]{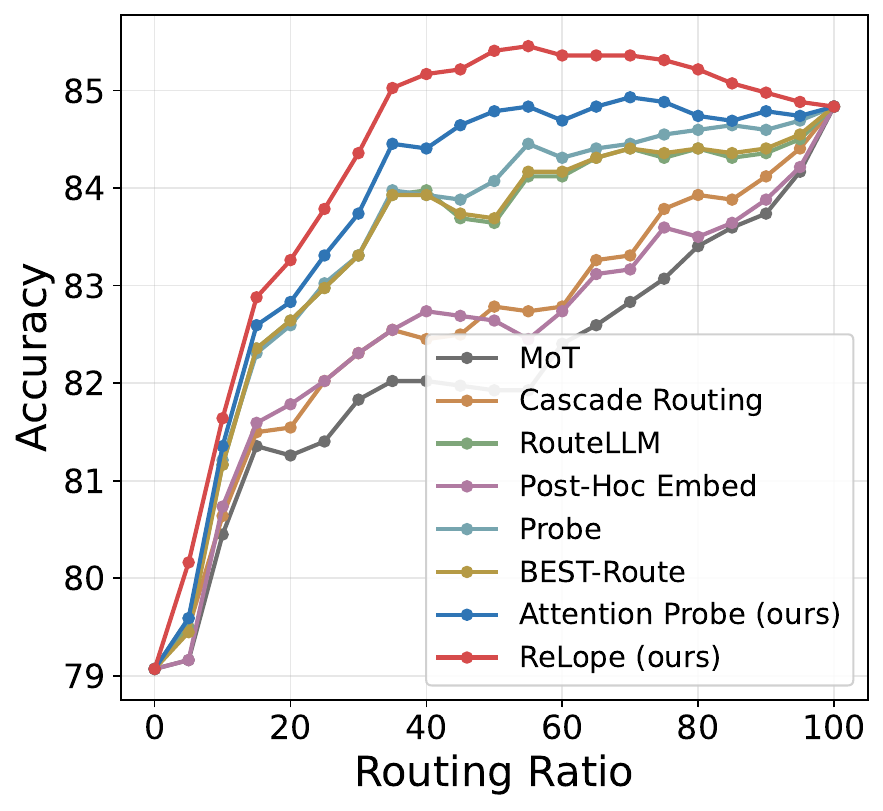}
\caption{ScienceQA}\end{subfigure}\hfill
\begin{subfigure}[t]{.32\linewidth}\centering
\includegraphics[width=\linewidth]{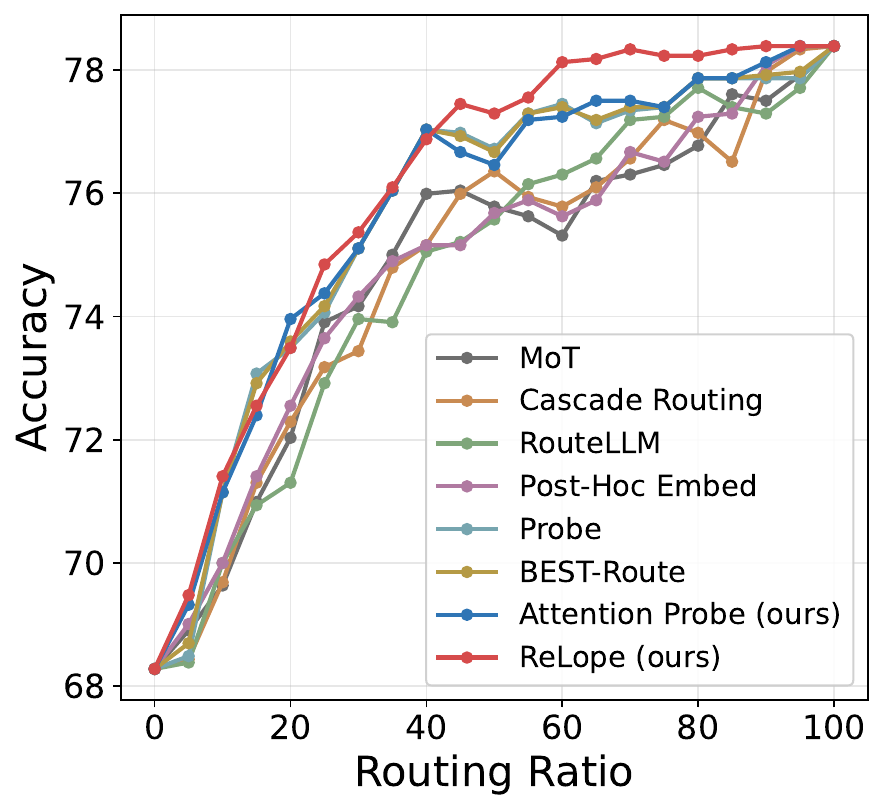}
\caption{ChartQA}\end{subfigure}
\caption{\textbf{Hybrid-system accuracy against escalation rate} with Qwen2.5-VL-7B-Instruct as the small model. At rate $h\%$ each router sends its $h\%$ least confident queries to the large model. The endpoints are the small model alone and the large model alone. A higher curve means more correct system answers for the same number of large-model calls.}
% Curves can cross because the large model makes mistakes too and can undo a correct small answer.
\label{fig:routing_ratio}
\end{figure}

A better prediction is only useful if it buys accuracy per large-model call. Figure~\ref{fig:routing_ratio} plots hybrid accuracy against escalation rate on A-OKVQA, ScienceQA, and ChartQA, using Qwen2.5-VL-7B-Instruct as the small model. The large model is GPT-4.1. Figure~\ref{fig:routing_ratio_extra} in Appendix~\ref{app:routing_extra} covers MMMU and MathVision. \method{} gives favorable trade-offs on all five tasks: for a given budget of large-model calls it returns more correct system answers than the other routers over most of the range. The curves do not follow AUC alone, since the large model can replace a correct small answer with a wrong one.
Appendix~\ref{app:cost} converts the same comparison from escalation rate into large-model cost.

The router's own overhead is small: with the backbone forward pass included, \method{} adds under two percent to probe's per-example time and about a tenth of a gigabyte of peak memory (Appendix~\ref{app:latency}). What grows is the training footprint, about thirty times more parameters than probe, and the next section asks whether the parameters, rather than the design, explain the gain.

\subsection{Where does the gain come from?}
\label{sec:ablation}
\begin{table}[t]
\begin{minipage}[t]{0.505\linewidth}
\caption{\textbf{Comparison in text-only and multimodal conditions.} Routing AUC (\%) on A-OKVQA under the two evidence conditions.}
\label{tab:readout_gap}
\centering\footnotesize\setlength{\tabcolsep}{2pt}
\begin{tabular}{lccc}
\toprule
Router & Original input & Caption only & $\Delta$ \\
\midrule
Probe & $81.94\pm0.58$ & $85.25\pm0.52$ & $+3.31$ \\
Attention Probe & $84.06\pm0.65$ & $85.31\pm0.47$ & $+1.25$ \\
\rowcolor{gray!15} ReLope & $86.24\pm0.63$ & $87.17\pm0.82$ & $+0.93$ \\
\bottomrule
\end{tabular}
\end{minipage}\hfill
\begin{minipage}[t]{0.465\linewidth}
\caption{\textbf{Routing scores under image perturbations.} AUC (\%) for predicting clean-input correctness on A-OKVQA.}
\label{tab:robust}
\centering\footnotesize\setlength{\tabcolsep}{3pt}
\resizebox{\linewidth}{!}{%
\begin{tabular}{lccccc}
\toprule
Method & Clean & Noise & JPEG & Blur & $\Delta$AUC$\downarrow$ \\
\midrule
Probe & 82.03 & 77.31 & 76.84 & 76.05 & 5.30 \\
Attention Probe & 84.24 & 82.11 & 81.65 & 80.90 & 2.69 \\
\rowcolor{gray!15} ReLope & \textbf{86.17} & \textbf{86.05} & \textbf{84.21} & \textbf{83.92} & \textbf{1.44} \\
\bottomrule
\end{tabular}}
\end{minipage}
\end{table}

\begin{table}[t]
\caption{\textbf{Capacity and adaptation controls} with Qwen2.5-VL-7B-Instruct (AUC, \%). LoRA + probe feeds the adapted last-token state to the classifier. VIB + probe applies the bottleneck to the frozen last-token state. Widened MLP enlarges the classifier to \method{}'s parameter count.}
\label{tab:rev_capacity}
\centering\small\setlength{\tabcolsep}{5pt}
\begin{tabular}{lrccc}
\toprule
Method & Trainable params & MMMU & A-OKVQA & ScienceQA \\
\midrule
Probe & 0.96M & $76.61\pm0.20$ & $81.94\pm0.58$ & $83.65\pm0.86$\\
LoRA + Probe & 3.84M & $77.92\pm0.51$ & $83.17\pm0.71$ & $85.25\pm0.73$\\
VIB + Probe & 26.66M & $76.82\pm0.87$ & $83.05\pm0.75$ & $84.44\pm1.05$\\
Widened MLP & 30.31M & $78.65\pm0.29$ & $83.87\pm0.21$ & $86.13\pm0.36$\\
\rowcolor{gray!15} ReLope & 29.54M & $\mathbf{80.79}\pm0.40$ & $\mathbf{86.24}\pm0.63$ & $\mathbf{91.78}\pm0.52$\\
\bottomrule
\end{tabular}
\end{table}

% Before the main-text sweep update:
% \textbf{Learned pooling narrows the image--caption gap.}
% Table~\ref{tab:readout_gap} shows that \attn{} improves AUC over the standard probe by 2.12 points with images but only 0.06 points with captions. The gap between the two conditions consequently falls from 3.31 to 1.25 points. Because the backbone remains frozen, this result shows that changing how its token hidden states are pooled makes correctness easier to predict under image input. It does not establish how much correctness information the frozen states contain.
\textbf{Learned pooling narrows the image--caption gap.}
To examine whether the fixed probe input limits correctness prediction, we compare the standard probe with \attn{} while keeping the backbone frozen. \attn{} improves AUC by 2.12 points with images but only 0.06 with captions, narrowing the gap from 3.31 to 1.25 points (Table~\ref{tab:readout_gap}). This suggests that how token hidden states are combined contributes substantially to the observed gap.

\begin{table}[t]
\centering
\begin{minipage}[t]{0.43\linewidth}
\caption{\textbf{The probe uses the best tested fixed hidden state.} A-OKVQA AUC (\%), Qwen2.5-VL-7B-Instruct.}
\label{tab:layer_token}
\centering\footnotesize\setlength{\tabcolsep}{3pt}
\begin{tabular}{rcccc}
\toprule
& Last & \multicolumn{3}{c}{Mean over} \\
\cmidrule(lr){3-5}
Layer & token & Text & Image & All \\
\midrule
4  & 64.19 & 67.26 & 56.23 & 56.87 \\
8  & 66.33 & 69.06 & 62.46 & 63.22 \\
12 & 68.28 & 70.08 & 66.95 & 67.17 \\
16 & 71.88 & 70.65 & 70.77 & 71.04 \\
20 & 81.84 & 77.20 & 74.33 & 75.54 \\
24 & \textbf{82.03} & 76.40 & 71.82 & 72.91 \\
28 & 80.07 & 71.92 & 69.20 & 69.13 \\
\bottomrule
\end{tabular}
\end{minipage}\hfill
\begin{minipage}[t]{0.55\linewidth}
\caption{\textbf{KL and LoRA rank sweeps.} AUC (\%) on A-OKVQA (Qwen2.5-VL-7B-Instruct). Counts cover trainable LoRA parameters; bold marks the best mean.}
\label{tab:hyperparameters}
\centering\footnotesize\setlength{\tabcolsep}{2.5pt}
\revfit{\begin{tabular}{cc@{\hspace{3mm}}crc}
\toprule
\multicolumn{2}{c}{KL coefficient} & \multicolumn{3}{c}{LoRA rank} \\
\cmidrule(lr){1-2}\cmidrule(lr){3-5}
$\beta$ & AUC (\%) & $r$ & LoRA params. & AUC (\%) \\
\midrule
$0$       & $84.08 \pm 0.26$ & 16  & 360{,}448     & $85.89 \pm 0.27$ \\
$10^{-3}$ & $84.18 \pm 0.61$ & 64  & 1{,}441{,}792 & $85.37 \pm 0.33$ \\
$10^{-2}$ & $85.17 \pm 0.35$ & 128 & 2{,}883{,}584 & $\mathbf{86.24} \pm 0.63$ \\
$10^{-1}$ & $86.06 \pm 0.26$ & 256 & 5{,}767{,}168 & $85.18 \pm 0.48$ \\
$1$       & $\mathbf{86.24} \pm 0.63$ & 512 & 11{,}534{,}336 & $85.05 \pm 0.72$ \\
\bottomrule
\end{tabular}}
\end{minipage}
\end{table}

\textbf{\method{} improves both conditions, with larger gains for images.}
Compared with the standard probe, \method{} increases AUC by 4.30 points with images and 1.92 points with captions, narrowing the image--caption gap to 0.93 points (Table~\ref{tab:readout_gap}). It also improves over \attn{} in both conditions, suggesting benefits from adapting and regularizing the routing hidden states beyond learned pooling. The larger gain with images is consistent with our focus on multimodal routing.

% Before the main-text sweep update:
% \textbf{Not capacity.}
% Widening the classifier to \method{}'s parameter count on frozen states recovers about two points over the probe and still trails \method{} by 2.14 to 5.65 points across the three datasets (Table~\ref{tab:rev_capacity}). Parameter count does not track AUC inside the table either, since VIB + probe carries seven times the parameters of LoRA + probe and gains less.
\textbf{Parameter count alone does not explain the gains.}
To test whether \method{} benefits mainly from additional parameters, we widen the classifier on frozen hidden states to a comparable parameter count. This control improves over the standard probe but still trails \method{} by 2.14--5.65 AUC points (Table~\ref{tab:rev_capacity}). VIB + Probe also uses more parameters than LoRA + Probe while gaining less. These comparisons suggest that how the parameters are used matters beyond their number.

% Before the main-text sweep update:
% \textbf{Adaptation and bottleneck together.}
% LoRA + Probe and VIB + Probe gain 1.4 and 0.7 AUC points on average over the standard probe (Table~\ref{tab:rev_capacity}). \method{} gains 4.18, 4.30, and 8.13 points on MMMU, A-OKVQA, and ScienceQA, outperforming both controls.
\textbf{Adaptation and bottleneck together.}
We evaluate LoRA + Probe and VIB + Probe separately to examine the contribution of each component (Table~\ref{tab:rev_capacity}). They improve AUC by 1.4 and 0.7 points on average over the standard probe, respectively. Combining them in \method{} outperforms both controls on all three datasets, with gains larger than the sum of the two individual gains. This pattern suggests complementary benefits from adapting and regularizing the routing hidden states.

% Before the main-text sweep update:
% \textbf{Layer and readout on frozen states.}
% Table~\ref{tab:layer_token} tests whether changing the layer or fixed readout can explain the gains, comparing four readouts at seven layers on A-OKVQA with the same classifier and one seed. The layer-24 last-token hidden state achieves the highest AUC (82.03); means over text, image, or all tokens do not improve on it. This supports the standard probe configuration used in our comparisons. Appendix~\ref{app:layer_token} gives the protocol.
\textbf{Layer and readout on frozen states.}
We test whether a different layer or fixed readout would provide a stronger standard probe, comparing four readouts at seven layers with the same classifier (Table~\ref{tab:layer_token}). The layer-24 last-token hidden state achieves the highest AUC, while means over text, image, or all tokens do not improve on it. Thus, our gains are measured against the strongest fixed configuration tested in this sweep. Appendix~\ref{app:layer_token} details the single-seed protocol.

% Before the main-text sweep update:
% \textbf{KL regularization and LoRA rank.}
% Table~\ref{tab:hyperparameters} shows that $\beta=1$ improves mean AUC by 2.16 points over $\beta=0$, with similar performance at $\beta=0.1$. Rank 128 achieves the highest mean AUC, while rank 512 uses four times as many LoRA parameters but performs worse on average. Thus, increasing rank does not consistently improve correctness prediction. Appendix~\ref{app:hyperparameters} details the sweep settings.
\textbf{KL regularization.}
To isolate the KL penalty, the $\beta=0$ condition retains the stochastic Gaussian bottleneck but removes the KL term. Adding the penalty at $\beta=1$ improves AUC by 2.16 (Table~\ref{tab:hyperparameters}), which supports the contribution of KL regularization beyond the stochastic bottleneck alone. 

\textbf{LoRA rank.}
We vary the adaptation capacity while keeping $\alpha/r$ and the adapted projections fixed (Table~\ref{tab:hyperparameters}). Rank 128 achieves the highest mean AUC, while increasing the rank to 512 uses four times as many LoRA parameters as rank 128 but yields lower mean AUC. Thus, greater adaptation capacity does not consistently improve correctness prediction.

\subsection{Robustness and transferability}
\label{sec:transfer}
\begin{table}[t]
\caption{\textbf{Cross-dataset transfer} with Qwen2.5-VL-7B-Instruct. ROC-AUC (\%) on each held-out dataset (column) for routers trained on the other four datasets. Bold marks the best in each column.}
\label{tab:ood}
\centering\small\setlength{\tabcolsep}{5pt}
\begin{tabular}{lccccc}
\toprule
\textbf{Method} & \textbf{MMMU} & \textbf{A-OKVQA} & \textbf{ScienceQA} & \textbf{ChartQA} & \textbf{MathVision} \\
\midrule
MoT & 67.52 & 72.35 & 72.08 & 75.38 & \textbf{82.59} \\
Cascade Routing & 67.15 & 72.94 & 75.33 & 65.47 & 63.28 \\
RouteLLM & 69.28 & 65.57 & 75.98 & 64.38 & 61.16 \\
Post-Hoc Embed & 67.55 & 66.18 & 69.81 & 65.75 & 64.39 \\
Probe & 71.78 & 81.92 & 80.57 & 73.05 & 66.35 \\
Attention Probe & 73.22 & 81.87 & 81.14 & 73.98 & 66.42 \\
\rowcolor{gray!15} ReLope & \textbf{76.66} & \textbf{83.68} & \textbf{81.58} & \textbf{75.57} & 66.83 \\
\bottomrule
\end{tabular}
\end{table}
\textbf{Image perturbations.}
One motivation for the bottleneck is to limit reliance on input-specific information. We therefore test whether each router can still distinguish originally correct from incorrect answers when the image appearance changes. We apply Gaussian noise, JPEG compression, and Gaussian blur to A-OKVQA images while keeping clean-input correctness labels fixed. As shown in Table~\ref{tab:robust}, the mean AUC drop across the three corruptions is 5.30 points for the probe, 2.69 for \attn{}, and 1.44 for \method{}. The smaller drop suggests that \method{} more stably predicts clean-input correctness under image perturbations. This fixed-label test does not assess the correctness of answers generated from corrupted images. 

\textbf{Held-out datasets.}
To test transfer, we hold out each dataset in turn, train each learned router on the other four, and report its AUC on the held-out dataset in the corresponding column of Table~\ref{tab:ood}. \method{} has the highest AUC on four of five targets, showing an advantage in cross-dataset transfer.  MathVision is the exception: its visual-math reasoning may pose a larger domain shift, and training-free MoT outperforms \method{} (82.59 versus 66.83 AUC). The results suggest that the routing features learned by \method{} capture correctness signals that transfer across diverse multimodal domains, rather than relying primarily on dataset-specific patterns.

\section{Conclusion}
\label{sec:conclusion}
Hidden-state probes provide an answer correctness signal for routing with little cost, but they rely on hidden states produced by models trained for answer generation. We find that correctness is harder to predict with image inputs than with captions, and that \attn{} substantially narrows this gap by learning to combine frozen token hidden states. 
Based on these findings, we propose \method{}, which learns hidden states for routing through a dedicated decision layer using a separate LoRA-adapted layer and a KL-regularized stochastic bottleneck, while preserving the answer-generation path. Across five benchmarks and three backbones, \method{} achieves the highest AUC among the evaluated methods in all 15 dataset--backbone pairs and improves hybrid accuracy--cost trade-offs with low inference overhead. These results support learning a dedicated decision layer within an MLLM to improve correctness prediction and routing accuracy--cost trade-offs.

\newpage

\section*{AI use statement}
We used generative AI tools to help rephrase the manuscript and check the writing errors. We have not used generative AI tools to formulate claims or proofs, or to implement the methods. The authors verified the technical content and are responsible for it.

\bibliography{custom,extra}
\bibliographystyle{iclr2027_conference}

\clearpage
\appendix
\section{Implementation Details}
\label{sec:implementation}
\subsection{Training configuration}
The current implementation uses PyTorch, Transformers, and PEFT. Probe-based routers use four Linear layers, $d\!\rightarrow\!256\!\rightarrow\!128\!\rightarrow\!64\!\rightarrow\!1$, with ReLU activations. We train for three epochs with AdamW, learning rate $10^{-4}$, and batch size 8. The official training pool is split by correctness label into 90\% train and 10\% development data; the checkpoint is selected only by development AUC and then evaluated once on the official validation split. The read layer is 24 for Qwen2.5-VL and Phi-4 and 40 for Gemma3. \method{} adapts the \texttt{q/k/v/o} projections of a separate copy of the read layer with LoRA rank 128, $\alpha=256$, and dropout 0.05. Its KL coefficient is $1.0$ with one-epoch linear annealing; KL is summed over latent dimensions and averaged over the batch. Training samples the bottleneck, while reported inference uses the posterior mean. Algorithm~\ref{alg:routing} lists the training steps.

\subsection{Evaluation protocol}
\label{sec:eval_protocol}
Correctness labels are obtained by comparing model answers with gold answers using dataset-specific evaluation rules (Appendix~\ref{sec:correctness_labels}). We evaluate correctness prediction using ROC-AUC, reporting the mean and standard deviation over three seeds unless otherwise stated. We assess routing utility through hybrid-system accuracy at different escalation rates and measure inference time and peak memory usage.

\subsection{Correctness labels}
\label{sec:correctness_labels}
For MMMU, ScienceQA, and A-OKVQA, the model output is normalized and its extracted option is compared with the gold option. ChartQA uses relaxed exact matching for numerical answers and normalized string matching for text answers. MathVision uses final-answer extraction and symbolic-equivalence checking. No judge model is used.

\subsection{Data splits and reproducibility}
Each current run stores the model revision, command-line configuration, immutable example IDs and split hashes, per-epoch development predictions, selected checkpoint, and final evaluation predictions. For Qwen2.5-VL on A-OKVQA, the official training pool yields 15,350 training and 1,706 development examples; final evaluation uses all 1,145 examples in the official validation split. Principal comparisons use seeds 42, 43, and 44; deviations are sample standard deviations across seeds.

Features are read at the last valid prompt token. Probe and Attention Probe may reuse cached frozen states. \method{} keeps the answer-generating backbone frozen and evaluates only the separately adapted routing layer after the shared lower layers, so matched routers score the same small-model answers and correctness labels. Single-seed analyses (Tables~\ref{tab:layer_token} and \ref{tab:robust}) are labeled as such and are not mixed with the three-seed results.

\subsection{Baseline inputs}
\label{app:baseline_inputs}
RouteLLM and Post-Hoc Embed both receive image information. RouteLLM is implemented as a similarity-weighted ranking router whose query representation concatenates a CLIP text embedding of the question and its options with a CLIP embedding of the image. The matrix-factorization variant uses the same representation. Post-Hoc Embed averages the final decoder layer of the small model over all non-padding tokens, so it reads the full multimodal context including image tokens, and it pools the last hidden layer rather than layer~24. These two rows therefore differ from the probe-based rows of Table~\ref{tab:auc_baselines} in both input and pooling.

MMR-MLP and MMR-CMR are adapted to binary correctness prediction using frozen OpenCLIP ViT-B/32 embeddings.

\section{Readout and Hyperparameter Analysis}
\label{app:readout_hyperparameters}
\subsection{Layer and token readout}
\label{app:layer_token}
Table~\ref{tab:layer_token} compares four frozen-state readouts at layers 4, 8, 12, 16, 20, 24, and 28 of Qwen2.5-VL-7B-Instruct on all 1{,}145 A-OKVQA validation examples. Each MLP probe follows Appendix~\ref{sec:implementation} and is trained with one seed. The layer-24 last-token hidden state gives the highest AUC among these configurations. Its AUC of 82.03 differs slightly from the three-seed mean of 81.94 in Table~\ref{tab:auc_baselines}.

\subsection{KL coefficient and LoRA rank}
\label{app:hyperparameters}
We separately sweep the KL coefficient $\beta$ and LoRA rank $r$ on A-OKVQA with Qwen2.5-VL-7B-Instruct; Table~\ref{tab:hyperparameters} in the main text reports the results. The $\beta=0$ condition retains the stochastic Gaussian bottleneck and removes only the KL penalty. In the rank sweep, we set $\alpha=2r$ to keep $\alpha/r$ constant and adapt the same \texttt{q/k/v/o} projections. Parameter counts include only trainable LoRA modules.

\section{Image--Caption Input Comparisons}
\label{app:e05}
\label{app:input_dependence}
\subsection{Comparison protocol}
Table~\ref{tab:motivation} uses all 1{,}145 A-OKVQA validation examples and the pinned Qwen2.5-VL-7B-Instruct revision. The same example IDs, answer parser, prompt template, and layer-24 last-token readout are used in every condition. Captions were generated by the pinned Qwen2.5-VL-7B-Instruct model with the prompt \emph{Describe this image in detail, focusing on visual facts needed to answer questions about it}. In the random-image condition the image is replaced by one drawn at random from another example of the dataset. Answers and correctness labels are recomputed per condition. Probe AUC follows the protocol of Table~\ref{tab:auc_baselines}: the MLP probe of Appendix~\ref{sec:implementation} is trained on the training pool of each condition, evaluated on the layer-24 last-token states of the validation examples, and reported as the mean and sample standard deviation over training seeds 42, 43, and 44. HSIC and label CKA are computed on the same states and labels. Caption input lowers answer accuracy by 10.05 points and raises probe AUC by 3.31 points, and HSIC and label CKA follow the same order. Because answers and labels are regenerated per condition, the comparison is descriptive rather than causal.

\subsection{Results on MMMU and ScienceQA}
We repeat the three input conditions of Table~\ref{tab:motivation} on MMMU and on the image subset of ScienceQA, keeping the pinned backbone, the caption prompt, the layer-24 last-token readout, and the per-condition retraining unchanged. Table~\ref{tab:probe_degradation} reports the result.

\begin{table}[!htb]
\caption{\textbf{The comparison of Table~\ref{tab:motivation} repeated on two further benchmarks.} Three input conditions on 900 MMMU and 2{,}097 ScienceQA validation examples with a frozen Qwen2.5-VL-7B-Instruct, under the protocol of Table~\ref{tab:motivation}. Probe AUC is the mean $\pm$ SD over three seeds. Bold marks the best value per column within each benchmark.}
\label{tab:probe_degradation}
\centering\small\setlength{\tabcolsep}{5pt}
\begin{tabular}{lcccc}
\toprule
Input & Acc.\ (\%) & Probe AUC & HSIC ($\times10^{-3}$) & Label CKA\\
\midrule
\multicolumn{5}{c}{\textit{MMMU}}\\
\midrule
Original input & \textbf{47.11} & $76.61\pm0.20$ & 4.332 & 0.1106  \\
Caption only & 38.56 & $\mathbf{78.02}\pm0.48$ & \textbf{4.969} & \textbf{0.1158}\\
Caption + random image & 40.89 & $75.74\pm0.64$ & 3.193 & 0.0904\\
\midrule
\multicolumn{5}{c}{\textit{ScienceQA}}\\
\midrule
Original input & \textbf{79.07} & $83.65\pm0.86$ & 3.183 & 0.1242\\
Caption only & 70.62 & $\mathbf{88.42}\pm0.03$ & \textbf{8.063} & \textbf{0.2151}\\
Caption + random image & 70.53 & $88.17\pm0.27$ & 7.188 & 0.2056\\
\bottomrule
\end{tabular}
\end{table}

On both benchmarks, replacing images with captions lowers answer accuracy but raises probe AUC, HSIC, and label CKA, reproducing the A-OKVQA pattern. Adding an unrelated image has little effect on ScienceQA but reduces probe AUC and both dependence measures on MMMU. Thus, the claim that evidence source matters more than image-token presence is benchmark-dependent. Section~\ref{sec:ablation} separately tests the role of the probe's readout.

\subsection{Illustrative input conditions}
\label{app:input_examples}

In Table~\ref{tab:motivation}, \emph{Input} names the prompt construction: the question and answer options, together with any image or caption. \emph{Evidence} identifies how information about the original scene is supplied, through the image or its generated caption. \emph{Image tokens} are visual tokens derived from an image supplied to the MLLM; caption words are text tokens.

\begin{table}[!htb]
\centering\small
\caption{\textbf{Concrete input constructions for the birthday question.} The question and answer options are unchanged across conditions. The cupcake photograph is an illustrative unrelated image, not a recorded random draw from the experiment.}
\label{tab:input_constructions}
\begin{tabular}{@{}p{0.23\linewidth}p{0.46\linewidth}cc@{}}
\toprule
Input condition & Content supplied with the question and options & Evidence & Image tokens \\
\midrule
Original input & The original birthday photograph, which shows the number 30 & image & \checkmark \\
Caption only & Its generated caption, which explicitly mentions 30; no image & text & -- \\
Caption + random image & The same birthday caption plus an unrelated photograph, e.g., the cupcake photograph & text & \checkmark \\
\bottomrule
\end{tabular}
\end{table}

Figure~\ref{fig:caption_examples} shows two concrete examples. For the cupcake question, the caption describes the topping as white frosting, supporting the gold answer, icing. For the birthday question, it explicitly states the number 30. Thus, the relevant clue is supplied visually in the original-input condition and textually in the caption-only condition. Adding an unrelated image introduces image tokens without restoring visual evidence from the original scene. These examples illustrate the input constructions, not the model's predictions or a guarantee that all captions preserve the required evidence.

\begin{figure}[!htb]
\centering
\begin{minipage}[t]{0.48\linewidth}
\centering
\includegraphics[height=2.8cm,width=\linewidth,keepaspectratio]{example_cake.jpg}
\par\smallskip
\raggedright\small
\textbf{(a) Cupcakes}\par
\textbf{ID:} {\footnotesize\texttt{2C8riXpRLX3CyM5jDz23m7}}\par
\textbf{Question:} What is the white substance on top of the cupcakes?\par
\textbf{Options:} A. butter; B. mayo; C. ice cream; D. icing.\par
\textbf{Gold answer:} D. icing.\par\smallskip
\textbf{Generated caption (201 tokens):}\par
\emph{The image shows the viewfinder of a camera displaying a close-up shot of three cupcakes being decorated. The cupcakes appear to be yellow cake with white frosting. Each cupcake is topped with a dollop of white frosting and a small red fruit, possibly a raspberry or strawberry, placed on top of the frosting. A hand is visible in the frame, holding what looks like a small tool or utensil, possibly used for placing the fruit toppings. The background within the camera's viewfinder includes a dark surface that appears to be a baking tray or a similar flat surface, and there is some powdered sugar scattered around. The camera's display also shows various details such as the battery level indicator, the number "100-0237" which likely refers to the file number, and the "CF" symbol indicating CompactFlash memory card usage. The camera model appears to be a Canon EOS 5D Mark II, as indicated by the branding at the bottom left corner of the viewfinder.}\par\smallskip
\textbf{Answer clue:} White frosting corresponds to icing.
\end{minipage}\hfill
\begin{minipage}[t]{0.48\linewidth}
\centering
\includegraphics[height=2.8cm,width=\linewidth,keepaspectratio]{example_birthday.jpg}
\par\smallskip
\raggedright\small
\textbf{(b) Birthday}\par
\textbf{ID:} {\footnotesize\texttt{2Aq5RiEn7eyfWjEbpuYT2o}}\par
\textbf{Question:} Which number birthday is probably being celebrated?\par
\textbf{Options:} A. one; B. ten; C. nine; D. thirty.\par
\textbf{Gold answer:} D. thirty.\par\smallskip
\textbf{Generated caption:}\par
\emph{The image shows a birthday cake placed on a wooden table. The cake is designed to look like a plush teddy bear, with a grayish-brown color and a soft texture that mimics the appearance of a stuffed toy. The teddy bear has a small blue bow tie around its neck and a tag attached to one of its paws, which reads "Happy Birthday."}\par\smallskip
\emph{To the right of the teddy bear-shaped cake, there is a smaller round cake with a purple base and a white top. The number "30" is prominently displayed on the top of this smaller cake, indicating that it might be part of a 30th birthday celebration. The smaller cake also has candles inserted into it.}\par\smallskip
\emph{The table is covered with a green tablecloth featuring a leafy pattern. In the background, there is a person sitting at the table, partially visible, wearing a red and black outfit. There are also some plates and glasses on the table, suggesting that the setting is a casual dining environment. The overall atmosphere appears festive and celebratory.}\par\smallskip
\textbf{Answer clue:} The caption explicitly identifies the number 30.
\end{minipage}
\caption{\textbf{Visual evidence and generated captions.} Two A-OKVQA examples with their images, questions, and full generated captions. Both captions retain the key clue needed for the corresponding question. Captions are reproduced verbatim. Gold answers are dataset annotations.}
\label{fig:caption_examples}
\end{figure}

\section{Hybrid-System Evaluation}
\label{app:hybrid_evaluation}
\subsection{Accuracy and cost at fixed escalation rates}
\label{app:cost}
\label{app:full_cost}
We evaluate fixed escalation rates on all 1{,}145 A-OKVQA validation examples by aligning cached GPT-4.1 answers and token counts with each router's scores (Table~\ref{tab:full_cost_saving}). At each rate, the lowest-scoring examples are escalated; rates are specified in advance without test labels. Accuracy is reported as mean $\pm$ SD over seeds 42, 43, and 44, and API cost is averaged across seeds. Costs cover strong-model API usage only; local overhead is reported separately in Appendix~\ref{app:latency}.

\begin{table}[!htb]
\caption{\textbf{Hybrid accuracy and large-model cost} on the A-OKVQA validation set with Qwen2.5-VL-7B-Instruct and GPT-4.1, at escalation rates fixed in advance. Accuracy is mean $\pm$ SD over three router seeds. Cost saved is relative to strong-only inference. Prices follow the published \texttt{gpt-4.1-2025-04-14} rates with no prompt-caching discount. Bold marks the best hybrid accuracy per budget.}
\label{tab:full_cost_saving}
\centering\small\setlength{\tabcolsep}{3.5pt}
\revfit{\begin{tabular}{lrrrrr}
\toprule
Router & Escalation & Hybrid acc. & GPT calls & API cost (\$) & Cost saved \\
\midrule
Small only & 0\% & 78.78 & 0 & 0.000 & 100.00\% \\
Strong only & 100\% & 90.48 & 1{,}145 & 1.098 & 0.00\% \\
\midrule
Probe & 20\% & $85.33\pm0.10$ & 229 & 0.214 & 80.54\% \\
Probe & 40\% & $88.68\pm0.25$ & 458 & 0.439 & 59.98\% \\
Probe & 60\% & $89.46\pm0.22$ & 687 & 0.654 & 40.45\% \\
\attn{} & 20\% & $\mathbf{85.59\pm0.46}$ & 229 & 0.220 & 80.00\% \\
\attn{} & 40\% & $89.14\pm0.13$ & 458 & 0.437 & 60.20\% \\
\attn{} & 60\% & $89.84\pm0.13$ & 687 & 0.656 & 40.24\% \\
\rowcolor{gray!15} \method{} & 20\% & $\mathbf{85.59\pm0.13}$ & 229 & 0.213 & 80.57\% \\
\rowcolor{gray!15} \method{} & 40\% & $\mathbf{89.32\pm0.20}$ & 458 & 0.437 & 60.16\% \\
\rowcolor{gray!15} \method{} & 60\% & $\mathbf{89.87\pm0.26}$ & 687 & 0.654 & 40.45\% \\
\bottomrule
\end{tabular}}
\end{table}

\method{} gives the highest mean accuracy at 40\% and 60\% escalation and ties \attn{} at 20\%. At 60\% escalation, it reaches 89.87\% accuracy versus 90.48\% for strong-only inference while saving 40.45\% of API cost, and at 40\% it is within 1.2 points of strong-only accuracy at 60.16\% cost saved. Appendix~\ref{app:calibration} complements this budget-matched comparison with fixed-threshold routing.

\subsection{Additional routing curves on MMMU and MathVision}
\label{app:routing_extra}
\begin{figure}[!htb]
\centering
\begin{subfigure}[t]{.42\linewidth}\centering
\includegraphics[width=\linewidth]{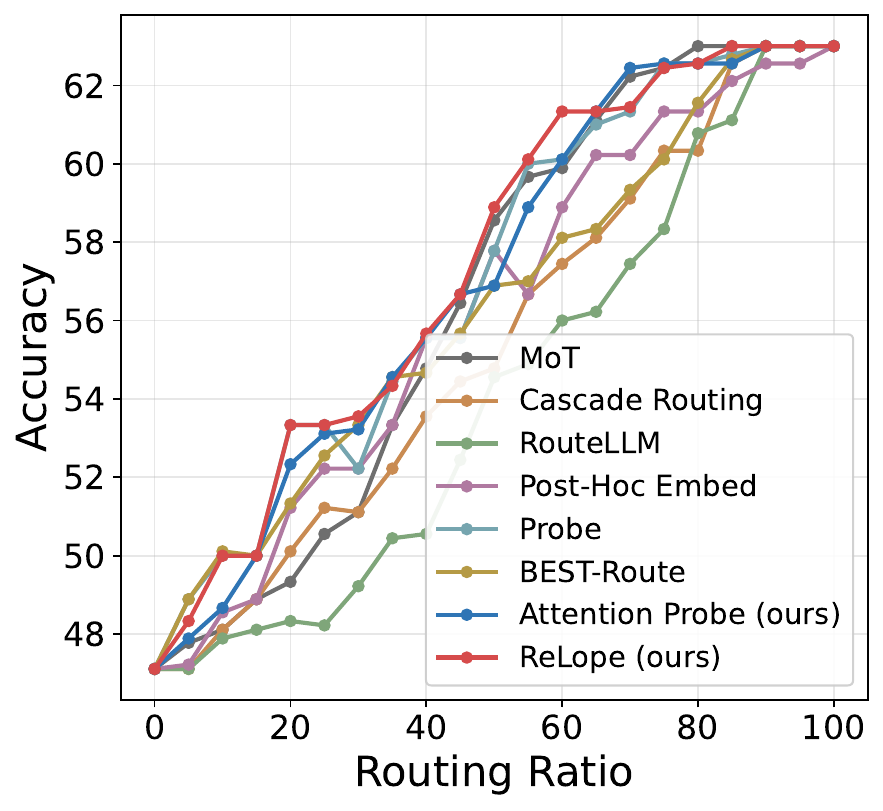}
\caption{MMMU}\end{subfigure}\hspace{.04\linewidth}
\begin{subfigure}[t]{.42\linewidth}\centering
\includegraphics[width=\linewidth]{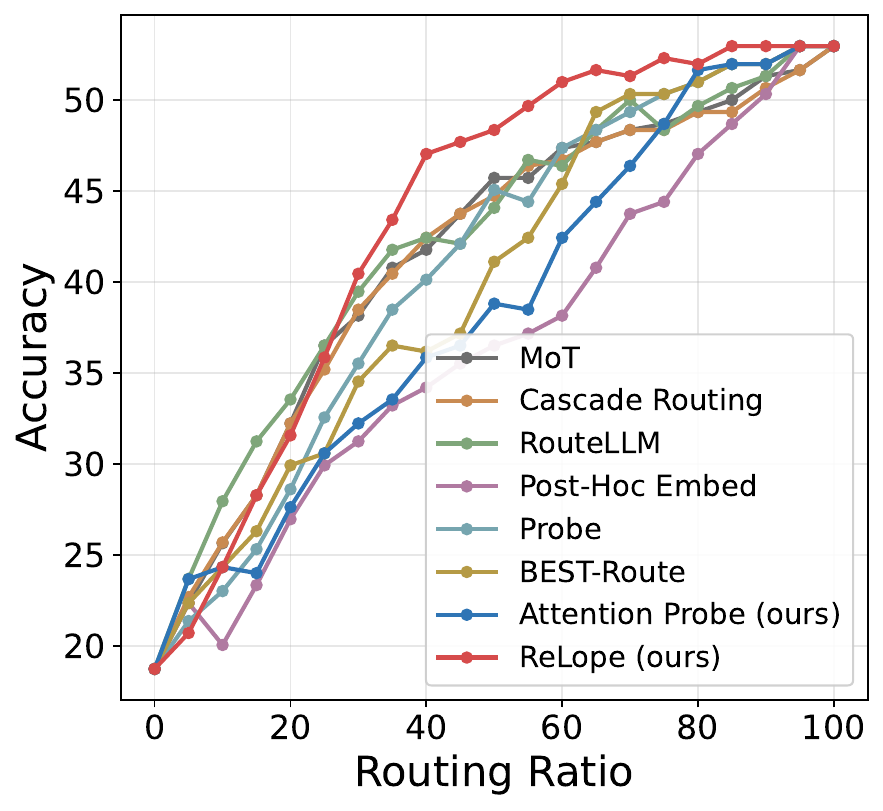}
\caption{MathVision}\end{subfigure}
\caption{\textbf{Hybrid-system accuracy against escalation rate} on the two benchmarks not shown in Figure~\ref{fig:routing_ratio}, with Qwen2.5-VL-7B-Instruct as the small model. The large model is GPT-4.1 on both MMMU and MathVision. Axes and endpoints follow Figure~\ref{fig:routing_ratio}.}
\label{fig:routing_ratio_extra}
\end{figure}

Both panels show the same pattern as Section~\ref{sec:system}. For a given budget of large-model calls \method{} returns at least as many correct system answers as the other routers over most of the range, and the curves cross in places because the large model can overturn a correct small-model answer.

\subsection{Calibration and threshold-based routing}
\label{app:calibration}
Fixed-threshold routing depends on score calibration as well as ranking. We evaluate ECE, Brier score, and escalation to GPT-4.1 when the routing score falls below $h$, following Appendix~\ref{sec:implementation} with seeds 42, 43, and 44. ECE uses 15 equal-width bins. Table~\ref{tab:calibration} reports hybrid accuracy at each threshold; because the same threshold escalates a different share of queries for each router, these accuracies are not budget-matched.

\begin{table}[!htb]
\caption{\textbf{Calibration and routing at a fixed score threshold.} A query is escalated when its routing score falls below $h$. Each cell gives the hybrid accuracy at that threshold, mean over three seeds; the share of queries escalated at a given $h$ differs across routers.}
\label{tab:calibration}
\centering\small\setlength{\tabcolsep}{4pt}
\revfit{\begin{tabular}{lrrrrrrr}
\toprule
Router & ECE & Brier & $h{=}0.5$ & $h{=}0.6$ & $h{=}0.7$ & $h{=}0.8$ & $h{=}0.9$\\
\midrule
\multicolumn{8}{c}{\textit{A-OKVQA} ($N=1{,}145$, small 78.78, large 90.48)}\\
\midrule
Probe & 0.0633 & 0.1257 & 82.47 & 85.15 & 86.90 & 87.98 & 89.34\\
\rowcolor{gray!15} \attn{} & 0.0480 & 0.1238 & 82.94 & 85.15 & 87.07 & 88.44 & 88.68\\
\rowcolor{gray!15} \method{} & \textbf{0.0409} & \textbf{0.1214} & 84.13 & 85.62 & 87.42 & 89.61 & 89.75\\
\midrule
\multicolumn{8}{c}{\textit{MMMU} ($N=900$, small 47.11, large 63.00)}\\
\midrule
Probe & 0.0948 & 0.2166 & 53.59 & 55.96 & 58.63 & 61.56 & 62.85\\
\rowcolor{gray!15} \attn{} & \textbf{0.0453} & 0.2385 & 54.48 & 59.85 & 63.00 & 63.00 & 63.00\\
\rowcolor{gray!15} \method{} & 0.0731 & \textbf{0.2130} & 54.85 & 58.30 & 60.78 & 62.59 & 63.00\\
\bottomrule
\end{tabular}}
\end{table}

\method{} has the lowest ECE and Brier score on A-OKVQA; on MMMU, \attn{} has the lowest ECE and \method{} the lowest Brier score. On MMMU, \attn{} matches the large-only accuracy of 63.00 at $h=0.7$. However, the same threshold can yield different escalation rates across routers, so these accuracies do not establish performance at a matched budget; Table~\ref{tab:full_cost_saving} provides that comparison.

\section{Computational Overhead}
\label{app:overhead}
\subsection{GPU inference overhead}
\label{app:latency}
We benchmark Qwen2.5-VL-7B-Instruct on ScienceQA on one NVIDIA H200 NVL (fp16), excluding preprocessing and host-to-device transfer but including CUDA synchronization. Head-only and backbone-plus-router timings average 200 and 30 repetitions, respectively. \method{} uses the posterior mean.

\begin{table}[!htb]
\caption{Local routing cost at batch size 4 on an NVIDIA H200 NVL (fp16). Router time covers the scoring heads; total time and throughput include the backbone and full routing computation.}
\label{tab:latency}
\centering\small\setlength{\tabcolsep}{4pt}
\begin{tabular}{lrrrr}
\toprule
Router & Head ms/sample & Total ms/sample & Samples/s & Trainable params \\
\midrule
Probe & 0.055 & 17.383 & 57.53 & 0.96M \\
Attention Probe & 0.047 & 17.372 & 57.56 & 0.96M \\
LoRA + Probe & 0.054 & 17.455 & 57.29 & 3.84M \\
\rowcolor{gray!15} ReLope & 0.065 & 17.695 & 56.51 & 29.54M \\
\bottomrule
\end{tabular}
\end{table}

At batch size 4, \method{} adds 0.31\,ms per example (1.79\%) over probe, with peak memory of 16.35 versus 16.24\,GiB. Differences remain small at batch sizes 1 and 8, but differing sequence lengths preclude a scaling comparison.

\subsection{CPU routing-branch cost}
\label{app:bs1}
To separate scoring-head cost from the full routing branch, we replay the adapted layer and scoring heads at batch size one on a memory-bandwidth-limited host CPU in float32, using the same inputs as in a forward pass (Table~\ref{tab:bs1_cost}).

\begin{table}[!htb]
\caption{\textbf{Routing cost at batch size one} on a host CPU (float32), separating the answer path, the routing branch, and the scoring heads.}
\label{tab:bs1_cost}
\centering\small\setlength{\tabcolsep}{4pt}
\revfit{\begin{tabular}{llrrrrrr}
\toprule
Dataset & Image size & Seq.\ len. & Answer path (ms) & Branch (ms) & Heads (ms) & Branch+heads / answer & Heads / branch \\
\midrule
ChartQA & native & 244 & 7{,}160.84 & 324.85 & 6.106 & 4.62\% & 1.88\% \\
ScienceQA & native & 104 & 3{,}953.45 & 239.76 & 4.569 & 6.18\% & 1.91\% \\
ScienceQA & 644 & 584 & 20{,}318.90 & 584.05 & 7.372 & 2.91\% & 1.26\% \\
ScienceQA & 896 & 1{,}079 & 43{,}154.09 & 1{,}110.66 & 4.742 & 2.58\% & 0.43\% \\
ScienceQA & 1288 & 2{,}171 & 118{,}573.52 & 2{,}359.37 & 4.322 & 1.99\% & 0.18\% \\
\bottomrule
\end{tabular}}
\end{table}

The branch plus heads cost 1.99--6.18\% of the answer path, with a smaller share for longer ScienceQA prompts. Heads account for under 2\% of branch time, distinguishing head-only timings from full-branch overhead.

\end{document}